\documentclass{article}
\usepackage{ijcai22}

\usepackage{times}
\usepackage{soul}
\usepackage{url}
\usepackage[hidelinks]{hyperref}
\usepackage[utf8]{inputenc}
\usepackage[small]{caption}
\usepackage{graphicx}
\usepackage{amsmath}
\usepackage{amsthm}
\usepackage{booktabs}
\usepackage{algorithm}
\usepackage{algorithmic}
\usepackage{subfigure}
\title{AggPose: Deep Aggregation Vision Transformer for Infant Pose Estimation}

\author{
Xu Cao$^{1,4}$\thanks{project lead}
\and
Xiaoye Li$^{1,2}$\thanks{contributed equally to the first-author} \and
Liya Ma$^{1,2}$ \and
Yi Huang$^{1,3}$ \and
Xuan Feng$^{1,3}$ \and
Zening Chen$^{4}$ \and \\
Hongwu Zeng$^{3}$ \And
Jianguo Cao$^{1,3}$\thanks{corresponding author}
\affiliations
$^1$Shenzhen Automatic Rehabilitation Laboratory \\
$^2$Shenzhen Baoan Women's and Childiren's Hospital, Jinan University \\
$^3$Shenzhen Children’s Hospital \\
$^4$New York University \\
\emails
xc2057@nyu.edu, caojgsz@126.com
}

\begin{document}

\maketitle

\begin{abstract}
 Movement and pose assessment of newborns lets experienced pediatricians predict neurodevelopmental disorders, allowing early intervention for related diseases. However, most of the newest AI approaches for human pose estimation methods focus on adults, lacking publicly benchmark for infant pose estimation. In this paper, we fill this gap by proposing infant pose dataset and Deep Aggregation Vision Transformer for human pose estimation, which introduces a fast trained full transformer framework without using convolution operations to extract features in the early stages. It generalizes Transformer + MLP to high-resolution deep layer aggregation within feature maps, thus enabling information fusion between different vision levels. We pre-train AggPose on COCO pose dataset and apply it on our newly released large-scale infant pose estimation dataset. The results show that AggPose could effectively learn the multi-scale features among different resolutions and significantly improve the performance of infant pose estimation. We show that AggPose outperforms hybrid model HRFormer and TokenPose in the infant pose estimation dataset. Moreover, our AggPose outperforms HRFormer by 0.8 AP on COCO val pose estimation on average.  Our code is available at \href{https://github.com/SZAR-LAB/AggPose}{github.com/SZAR-LAB/AggPose}.
\end{abstract}


\section{Introduction}

Each year, approximately 5 million newborns around the world are suffering from neurodevelopmental disorder. Due to the lack of early diagnosis and intervention, many infants are severely disabled and abandoned by their parents, especially in countries with limited numbers of pediatricians with extensive experience in neurodevelopmental disorders. This has become a conundrum that plagues many families around the world. 

Recent developments in deep learning based approaches open possibilities for developing computer-aid movement assessment tools in early intervention for neurodevelopmental disorder. One of the most predictive tools for early cerebral palsy diagnosis is general movements assessment (GMA), as it needs to discriminate fidgety from non-fidgety movements in many small-amplitude movements~\cite{silva2021future}, where computers are more sensitive to detect such movements. Researchers used human pose estimation methods like OpenPose~\cite{cao2019openpose} to capture infant pose and then generate infant motion sequence to detect cerebral palsy. Compared with manual GMA detection, computer-based approaches are much faster with low cost. However, this task is challenging in real applications considering complex scenarios for infant pose and there is a lack of large-scale public infant pose datasets around the world. Besides, the 17 adult keypoints defined by the COCO dataset do not support infant movement detection well due to the lack of clinical significance and actionability.

Another problem is the performance of the pose estimation methods. Although CNN-based methods have pushed human pose estimation to a new level thanks to the intense representation learning and semantics understanding ability, it is still not performing well to understand global constraint relationships between body parts~\cite{li2021tokenpose}. Researchers combined Vision Transformer with CNN into hybrid models to address this issue, let the ViT expand the receptive field, and enhance the model's ability to capture constraint relationships between body parts. Among recent advancements, the local-window self-attention structure from Swin Transformer~\cite{liu2021swin}, and Mix Feed Forward Network (Mix-FNN) from SegFormer~\cite{xie2021segformer} showed great potential in the direction of multi-scale feature representation learning~\cite{gu2021multi}.


However, some issues still make it challenging to apply Transformer for human pose estimation: (1) The first stages of the hybrid models highly rely on the pretrained HRNet convolutional layers, which can not utilize large-scale unlabeled data with newest self-supervised masked autoencoder~\cite{he2021masked}; (2) Hard to converge during the training process; (3) Models are challenging to transfer from one domain to another domain.

In this paper, we propose AggPose, a generalization of multi-scale transformer architecture to the deep aggregation network. Different from HRFormer, AggPose does not use convolutional layers for initial feature extractor and fusion modules. Instead, it uses layer-by-layer Mix Transformers and a cross-resolution MLP fusion module. The Transformer receives input from the former layer, applies self-attention operation and Mix-FNN, and sends the message to the next layer. The MLP fusion module integrates richer spatial information from different resolution levels and sends the result to the next stage.
We conduct experiments on COCO human pose estimation dataset and then fine-tune the model on our proposed large-scale labeled infant pose estimation dataset. AggPose achieves competitive performance on both benchmarks. For example, AggPose-L gains 0.8 AP and 0.6 AP over HRFormer and TokenPose on COCO val set. AggPose-L's robustness and fast convergence make it easy to transfer from the COCO dataset to our infant pose dataset and achieve the highest 95.0 AP.

The contributions are summarized as follows:

\begin{itemize}
\item We propose a new Transformer + MLP based aggregation architecture without using HRNet's CNN backbone and CNN-based multi-scale fusion modules.
\item To enhance the efficiency of deep layer aggregation, we design a special deep aggregation MLP structure to fuse information across different resolutions.
\item To facilitate research in early intervention for neurodevelopmental disorders, we present a large-scale infant challenging dataset including 20,748 pose labeled images. Experimental results show our framework's robustness in both COCO and this new dataset. To the best of our knowledge, this is the largest dataset constructed for infant pose estimation for clinical application.
\end{itemize}

\section{Related Works}

\subsection{Infant Pose Estimation}

Infant pose estimation has been found to have high application value in clinical research. Most commonly used cerebral palsy assessment tools such as GMA and CPVC~\cite{abbruzzese2020assessing} are already using automatic pose estimation methods to aid diagnosis. However, most existing algorithms are based on traditional machine learning methods for automatic infant pose estimation, limiting their capability to deal with complex conditions~\cite{silva2021future}. Meanwhile, there are very few attempts initiated by the artificial intelligence community to handle infant images. Only \cite{mccay2019establishing,reich2021novel} gave primacy attempts to adopt OpenPose~\cite{cao2019openpose} to extract key-points for infants but lack a large and general dataset. MINI-RGBD~\cite{hesse2018computer} is the most famous open-source dataset in this field, where it only contains 700 authentic infant images and a small set of synthesized infant images. All of these make automatic infant pose assessment methods unreliable in the real world clinical systems.

\begin{figure}[h]
\centering
\subfigure[] 
{
	\centering          
	\includegraphics[width=0.4\linewidth]{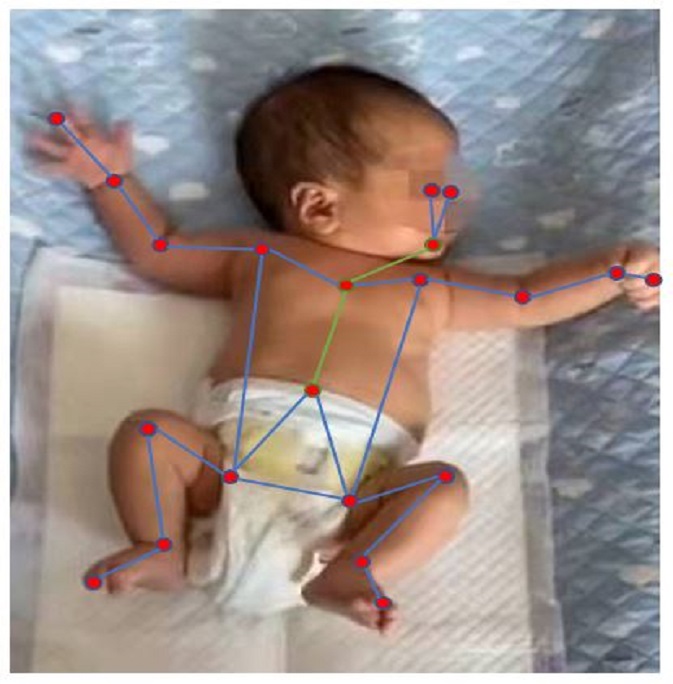}
}
\subfigure[]
{
	\centering     
	\includegraphics[width=0.4\linewidth]{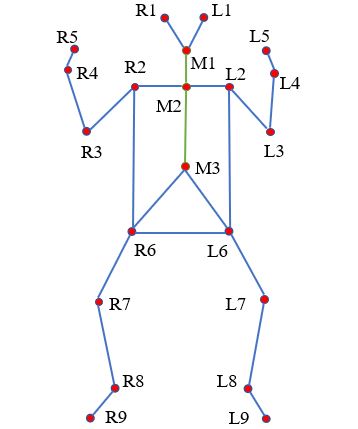}  
}
\caption{(a) Examples for our proposed InfantPose dataset. (b) 21 infant body key-points}
\label{fig:dataset_example}
\end{figure}

\subsection{Vision Transformers for Pose Estimation}
For many years, deep convolutional neural networks have been applied to human pose estimation. Among all CNN-based pose estimation algorithms, the schemes that maintain high-resolution representations throughout the network achieved great success. The most representative models are HRNet~\cite{wang2020deep}, HigherHRNet~\cite{cheng2020higherhrnet}, UDP~\cite{huang2020devil}, DARK~\cite{zhang2020distribution}. However, it is still tricky for CNNs to capture constraint relationships between human keypoints, as CNN's receptive field restrict its ability to understand global spatial relationships.

Recent several works have introduced Transformer for human pose estimation~\cite{yuan2021hrformer,yang2021transpose,li2021tokenpose}. TokenPose~\cite{li2021tokenpose} introduced Transformer with representing key-points as token embeddings for human Pose estimation. HRFormer~\cite{yuan2021hrformer} integrated HRNet with Swin Transformer~\cite{liu2021swin}, which makes full use of multi-resolution parallel information over different non-overlapping image windows. However, both HRFormer and TokenPose did not discard convolution operations to obtain initial features, as the first stage of HRFormer and TokenPose were fine-tuned on HRNet's CNN backbone. In this work, we propose Aggregation Vision Transformers (AViT), which provides a different way to solve the low-resolution problem of ViT and replace convolution operations with overlapping patch embedding to extract features in the early stages.

\section{Proposed Method}

\begin{figure*}[h]
  \centering
  \includegraphics[width=1.0\linewidth]{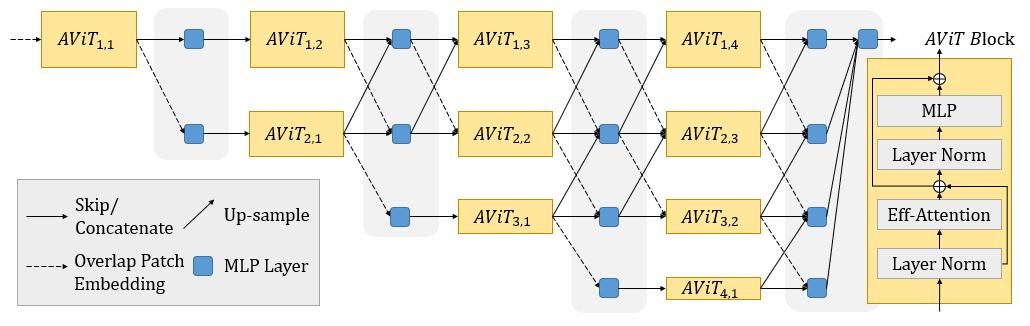} 
  \caption{The proposed AggPose architecture. Each module consists of multiple successive Mix Transformer blocks. Features across different resolutions are connected by MLP layer (blue square in the figure).}
  \label{fig:architecture}
\end{figure*}

Our goal is to propose a new infant pose dataset and build a new benchmark that can fast extract infant pose via vision transformers. Figure~\ref{fig:architecture} shows the pipeline of the model. Our infant pose estimation research have passed the ethics checks of Shenzhen Baoan Women's and Childiren's Hospital.

\subsection{Infant Pose Detection Dataset}

In this paper, we present a large-scale challenging dataset for newborn pose extraction and detection. It can be applied to predict infant movement sequence and design automatic clinical tools like automatic GMA. Despite the importance and difficulty of infant pose detection, existing datasets are either too small or too simple, and a large public annotated benchmark is needed to compare different methods. Besides, none of these datasets proposed suitable keypoints annotation for infant images, as they adopt the COCO's 17 keypoints format, while it loses many significant refined pose and movement features for the infant.

\begin{table}
\centering
\begin{tabular}{lrrrr}
\toprule
Dataset  & Videos & Labeled Images & Unlabeled \\
\midrule
MINI RGBD    & 12   & 700  & -       \\
COCO (infant)    &  0  & 1904  &  -      \\
SyRIP    & 0  & 1700  & -    \\
Ours    & 5187  & 20748  & 15 million    \\
\bottomrule
\end{tabular}
\caption{Comparison between other infant pose dataset.}
\label{tab:dataset}
\end{table}

Inspired by ~\cite{silva2021future,huang2021invariant}, we publish our new open-source infant pose dataset and new infant keypoints format. To collect data, we adopt GMA devices to record infant movement videos from 2013 to now. More than 216 hours of videos were collected, and 15 million frames were extracted. Both the size and the scalability of our dataset are much better than the MINI-RGBD dataset~\cite{hesse2018computer}. We randomly sampled 20,748 frames from the videos and let professional clinicians annotate infant keypoints. Then, we divided the dataset into 11,756 for the training set and validation set, 8,992 for the test set. The 21 keypoints format for infant pose is proposed by experienced clinicians who have researched neurodevelopmental disorders over 30 years.Figure~\ref{fig:dataset_example} shows some examples, considering clinical application requirements and protection of patient's privacy, our dataset reduces keypoints on infants' heads and comprises more refined body keypoints like fingers, toes, and navel. For public version, we will reformat the dataset to solve ethical issues: all infants' heads will be covered with mosaics in the final published keypoint dataset to preserve the patients' privacy. Commercial usage of infant pose dataset is prohibited.


In this paper, we focus on the human/infant supine position pose detection, which is the most straightforward application for the new presented dataset. However, this dataset can also be used in other clinical fields, as it contains over 200 hours of infant movement sequence and has a high relationship with the automated prediction of cerebral palsy and other  neurodevelopmental disorders. We hope applying our dataset and AggPose to early diagnosis and intervene disorders, promoting well-being for all at all ages, especially the children. In the future, we will also release more than 200 hours of new infant pose sequences generated from AggPose, and associated GMA labels. The retrospective study was approved by our institutional review board.

\subsection{Deep Aggregation Vision Transformers}

\subsubsection{Overlapped Patch Embedding} 
Early convolutions were considered practical tools to extract low-level features for hybrid transformer architectures. It is due to that transformers in the early stage treat the input as 1D vectors and exclusively focus on modeling the global context, which lose detailed localization information. HRNet and its pre-trained CNN parameters are the cornerstones of almost all the latest models for human pose estimation. Inspired by SegFormer~\cite{xie2021segformer}, we adopt full Transformer with Overlapped Patch Embedding to replace HRNet's CNN feature extractor and down-sampling stem of each stage. Compared with the early convolutions in HRNet, HRFormer, and TokenPose, Overlapped Patch Embedding can obtain better low-level features, enhancing the high-resolution Transformer's feature representation, and reduce computation complexity.

\subsubsection{Aggregation Vision Transformers (AViTs) Architecture} 
We follow the transformer module design from Mix Transformer~\cite{xie2021segformer} and start from high-resolution feature maps generated by the overlapped patch embedding with patch size = 7, stride = 4, and padding size = 3 as the first stage. Then, we add high-to-low resolution streams one by one via overlapped patch embedding. We use multiple multi-head self-attention blocks for each resolution stream to update feature representation. To construct different depth of models, we propose small (AggPose-S), and large (AggPose-L) model, respectively. Table~\ref{tab:transformer_blocks} shows the number of transformer layers for each stage in AggPose-L.

Compared with Swin Transformer and HRFormer, we do not use local-window self-attention to augment local information understanding considering the usage of overlapped patches. Instead, we use the sequence reduction process refer to \cite{xie2021segformer} and \cite{wang2021pyramid}, which significantly reduces the amount of calculation inside the transformer and accelerates the convergence process during model training. For each Transformer block, the self-attention is estimated as:

\begin{align}
    K = Linear(\gamma C, C)(K.Reshape(\frac{N}{\gamma}, \gamma C)) \\
    Attention(Q, K, V) = Softmax(\frac{QK^{T}}{\sqrt{d_{head}}})V
\end{align}

,where K is the token representation with initial shape $N \times C$. $\gamma$ is the reduction ratio that decrease the dimension of K from $N \times C$ to $N/\gamma \times C$.


\begin{table}
\centering
\begin{tabular}{lrrrr}
\toprule
Features level  & Stage 1 & Stage 2 & Stage 3 & Stage 4 \\
\midrule
1/4 & 3  & 3    & 3 & 3      \\
1/8 &   & 6 & 3 & 3       \\
1/16    &   &   & 40    & 3      \\
1/32    &   &   &   & 3      \\
\bottomrule
\end{tabular}
\caption{The number of Transformer layers for each stage.}
\label{tab:transformer_blocks}
\end{table}

\subsubsection{MLP Cross-Layer Aggregation}

Both ViT and Swin Transformer uses positional embedding to introduce the location information across layers. However, the resolution of positional embedding is fixed. For the local-window Transformer, there is lacking information exchange across the windows. Thus, both SegFormer and HRFormer introduced $3 \times 3$ depth-wise convolution into the feed-forward network (FFN) to expand the receptive field and reduce the harmful effect caused by positional embedding. The FFN with depth-wise convolution (HRFormer) and Mix-FFN (SegFormer) used a very similar calculation:

\begin{align}
    y = MLP(Activation(DWConv(MLP(x)))) + x
\end{align}

where $DWConv$ is a $3 \times 3$ depth-wise convolution operation.

In AggPose, we expand the usage of Mix-FFN into the deep aggregation approach across different resolution layers. For deep aggregation in CNN such as CAggNet~\cite{cao2021caggnet} and HRNet, the aggregation begins at the shallowest, high resolution layer and then iteratively merges deeper, low resolution layer. In this way, shallow features are refined as they are propagated through different stages of aggregation. Related research showed that deep aggregation structure propagates the aggregation of all resolutions instead of the preceding block alone to better preserve features. It is widely used for semantic segmentation tasks and to achieve competitive performance. In our work, the proposed cross-layer aggregation module consists of two main steps for each resolution level. First, multi-level features from different resolutions go through a mixed feed-forward network with $3 \times 3$ depth-wise convolution to unify the channel dimension and upsample or downsample (overlapped patch embedding) the feature map to the same shape. Then, we concatenate the feature vector from adjacent levels together and adopt an additional FFN layer to fuse the cross-layer information. Compared to convolutional multi-scale fusion modules in HRFormer and HRNet, MLP fusion modules accelerate convergence while improving model performance.

\begin{align}
x_{i,j}=\left\{
\begin{array}{rcl}
OverlappedPE_{i,j}(FFN(x_{i}))  &      & i < j  \\
x_{i}   &      & i=j   \\ 
Upsample_{i,j}(FFN(x_{i}))  &      & i > j
\end{array} \right.
\end{align}

where $x_{i,j}$ is the input of aggregation MLP layer. $x_{i}$ denotes the feature map from adjacent resolution. The cross-layer aggregation module is defined as

\begin{align}
    x_{j}= MixFFN(Concat(x_{j-1}, x_{j}, x_{j+1})) + x_{j}
\end{align}

where $MixFFN$ represents the Mix feed forward block in formula (3).

\subsection{Analysis}

There are two main benefits of AggPose and our large-scale infant pose dataset over other CNN or hybrid CNN Transformer methods like HRFormer and TokenPose and other small dataset for infant pose estimation, which are concluded as follows.

\paragraph{(1) Potential of using self-supervised learning.} Recently, Vision Transformers pre-trained with self-supervised learning have attracted much attention. MAE~\cite{he2021masked} construct an inpainting masked autoencoder task to learn representation from unlabeled data and fine-tuning the model on any supervised tasks. Their results prove that full transformers can learn reasonable semantic from large-scale unlabeled dataset. As Table 1 shows, we have plenty of unlabeled infant movement frames from 5,187 videos. All these data would be helpful for pre-training transformer-based autoencoder via self-supervised learning.

\paragraph{(2) Faster convergence.} In HRFormer, the feature passing is achieved via cross-layer convolution operation, which, is difficult to convergence. In our AggPose framework, messages are propagated by MLP across different layers. It can be viewed as a kind of modification to the deep layer aggregation model. As our experiments will show, such message pass scheme achieves better results than hybrid CNN-Transformer based methods.

\begin{table*}
\centering
\begin{tabular}{lrrrrrrrrr}
\toprule
Method    & Input size  & Backbone    & GFLOPs    & $AP$  & $AP^{50}$ & $AP^{75}$ & $AP^{M}$ & $AP^{L}$   & $AR$ \\
\midrule
SimpleBaseline-Res152   & 256$\times$192    & -  & 15.7  & 72.0  &  89.3  & 79.8  &  68.7  & 78.9  & 77.8   \\
HRNet-W32~\cite{wang2020deep}   & 256$\times$192    & -  & 7.1  & 74.4  &  90.5  &  81.9  & 70.8   & 81.0  &  79.8   \\
HRNet-W48~\cite{wang2020deep}   & 256$\times$192    & -  & 16.0  &  75.1  &  90.6  &  82.2  &  71.5  &  81.8  &  80.4     \\
TransPose~\cite{yang2021transpose}  & 256$\times$192  & HRNet   & 21.8    & 75.8 & 90.1  & 82.1  & 71.9  & 82.8  & 80.8      \\
TokenPose-L/D24~\cite{li2021tokenpose}  & 256$\times$192    & HRNet  & 11.0  & 75.8  & 90.3 &  82.5  & 72.3  &  82.7  & 80.9   \\
HRFormer-B~\cite{yuan2021hrformer}  & 256$\times$192    & HRNet  & 12.2  & 75.6  & \textbf{90.8}  & 82.8  & 71.7  & 82.6  & 80.8    \\
AggPose-S  & 256$\times$192 & MiT-B2  & 9.0   & 75.2  & 89.9  &  82.0  &  71.4  &  82.4   &  80.3     \\
AggPose-L  & 256$\times$192 & MiT-B5  & 15.0    & \textbf{76.4}  & 90.6  & \textbf{82.9}  & \textbf{72.7}   & \textbf{83.4}   & \textbf{81.3}\\
\bottomrule
\end{tabular}
\caption{Comparisons on the COCO validation set, provided with the same detected human boxes from HRNet.}
\label{tab:COCO_comparison}
\end{table*}

\section{Experiment}

\subsection{Model Variants}

Considering that the training process of most Transformer-based pose estimation models is complicated, we provide an effective training policy in this paper. First, we load the Mix Transformer~\cite{xie2021segformer} pre-trained on ImageNet, training Mix Transformer on the COCO keypoints training set. After the Mix Transformer converges, we load the parameter of Mix Transformer into each layer of AggPose. Then, we fixed the parameters of AggPose at different resolution levels layer by layer and fine-tuned the model on COCO and infant pose dataset. 

The configuration details for the size of overlapped patch embedding and the number of transformer layers are presented in Table~\ref{tab:transformer_blocks}. Note, Table~\ref{tab:transformer_blocks} only provide the configuration for AggPose-L, which uses MiT-B5 as the backbone. For AggPose-S, we used MiT-B2 as backbone, the number of transformer layers is [[3,3,3,3],[4,3,3],[6,3],[3]].

\subsection{Comparing with SOTA Methods}

\paragraph{Dataset.} We study the performance of AggPose on the COCO human pose estimation dataset~\cite{lin2014microsoft}, which contains more than 250K person instances labeled with 17 keypoints, and the new infant pose estimation dataset, which contains 20k infant instances labeled with 21 keypoints. MPII dataset is not used in our experiment due to its size (25K) is much smaller than COCO and has different keypoints format.

\paragraph{Training setting.} Following most of the default training and evaluation settings from HRNet and HRFormer, we trained the models using AdamW optimizer and an initial value of 0.001 as the learning rate. For the training batch size, we chose 32 due to limited GPU memory. The experiment takes 4 $\times$ 48G-RTX8000 GPUs. We follow the data augmentation in ~\cite{wang2020deep} mainly. 

\paragraph{Evaluation metric.} For COCO dataset, we adopt the default standard average precision (AP) as our evaluation metric. AP is calculated based on Object Keypoint Similarity (OKS):

\begin{align}
    OKS = \frac{\sum_{i}exp(-\frac{\hat{d}_{i}^{2}}{2s^{2}k_{i}^{2}})\delta(v_{i} > 0)}{\sum_{i}\delta(v_{i} > 0)}
\end{align}

where $\hat{d_i}$ is the L2 distance between the i-th keypoint and the groundtruth. $v_{i}$ denotes the visibility of the keypoint. $k_i$ is a keypoint-specific constant, which is different for different keypoint. We adopt the same evaluation metric to COCO for the infant pose dataset. As the new proposed infant pose has 21 keypoints, we set $k_i$ of each keypoint to the same value.

\paragraph{Keypoints detection on COCO pose estimation.} Table~\ref{tab:COCO_comparison} shows the comparisons on COCO val set. We compare AggPose with several state-of-art methods, including HRFormer~\cite{yuan2021hrformer}, TokenPose~\cite{li2021tokenpose}, TransPose~\cite{yang2021transpose}, HRNet~\cite{wang2020deep}. For input size of 256$\times$192, AggPose-L achieves 76.4 AP, which is best among all methods. We believe that AggPose-L can achieve better results after applying the newest distribution-aware coordinate representation~\cite{zhang2020distribution} or UDP~\cite{huang2020devil}. AggPose-L achieves 75.7 AP on the COCO test-dev set with 256 $\times$ 192 input size. 

\paragraph{Keypoints detection on infant pose estimation.} Table~\ref{tab:infant_comparison} reports the comparisons on our infant pose test set. We compare AggPose to the most representative bottom-up method OpenPose, as it is used by almost all newest proposed infant pose estimation frameworks~\cite{silva2021future}. We also compare AggPose to several recent CNN and hybrid Transformer models such as HRNet, TokenPose, and HRFormer. AggPose gains the highest 95.0 AP with an input size of 256$\times$192. During the training, we also find that both AggPose and HRNet perform better and converge faster than hybrid model such as TokenPose, and HRFormer. Though all of these models are pre-trained on COCO dataset, full CNN or full Transformer based methods are more robust after we fine-tune them on other domain like infant pose data.

\begin{table}
\centering
\begin{tabular}{lrrrr}
\toprule
Method  & image size  & AP  & AR\\
\midrule
OpenPose          & 256$\times$192  & 90.2  & 91.1      \\
SimpleBaseline-Res152  & 256$\times$192  & 93.9  & 94.9      \\
HRNet-W48         & 256$\times$192  & 94.5    & 95.6    \\
HRFormer-B        & 256$\times$192  & 93.8    & 95.0    \\
TokenPose-L/D24   & 256$\times$192  & 93.0     & 93.9    \\
AggPose-L         & 256$\times$192  & \textbf{95.0}     & \textbf{95.7}    \\
\bottomrule
\end{tabular}
\caption{Comparisons on the infant pose test set, provided with the same object detection boxes (OpenPose do not need object detection, as it is a bottom-up method. We select OpenPose for comparison is because most of newest proposed infant pose estimation frameworks are choose OpenPose as backbone.)}
\label{tab:infant_comparison}
\end{table}

\begin{table}
\centering
\begin{tabular}{lrrrr}
\toprule
Method  & image size & GFLOPs & AP \\
\midrule
Swin-B       & 256$\times$192  & 17.6  & 74.3      \\
SegFormer-B5 & 256$\times$192  & 12.3     & 74.2       \\
AggPose-L    & 256$\times$192  & 15.0     & \textbf{76.4}    \\
\bottomrule
\end{tabular}
\caption{Comparisons to Non-aggregation framework (Swin Transformer, SegFormer) on COCO pose estimation val}
\label{tab:abulation_comparison}
\end{table}

\begin{figure*}[h]
  \centering
  \includegraphics[width=1.0\linewidth]{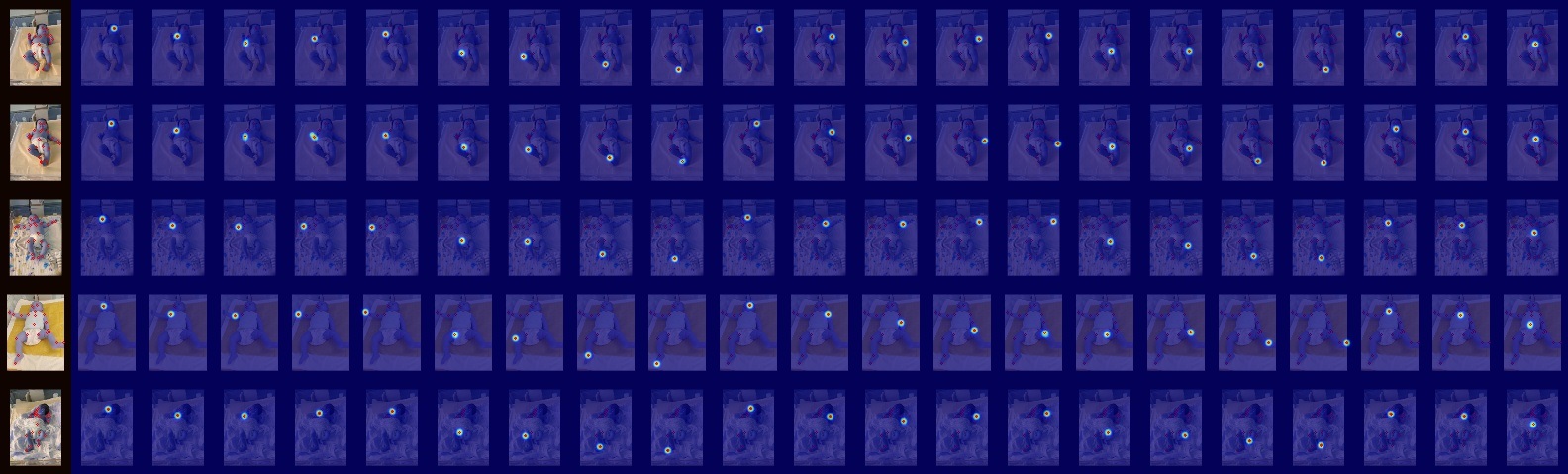} 
  \caption{Visualization of the pose estimation heatmap results based on AggPose-L on infant pose test set.}
  \label{fig:heatmap_infant}
\end{figure*}

\begin{figure}[h]
  \centering
  \includegraphics[width=1.0\linewidth]{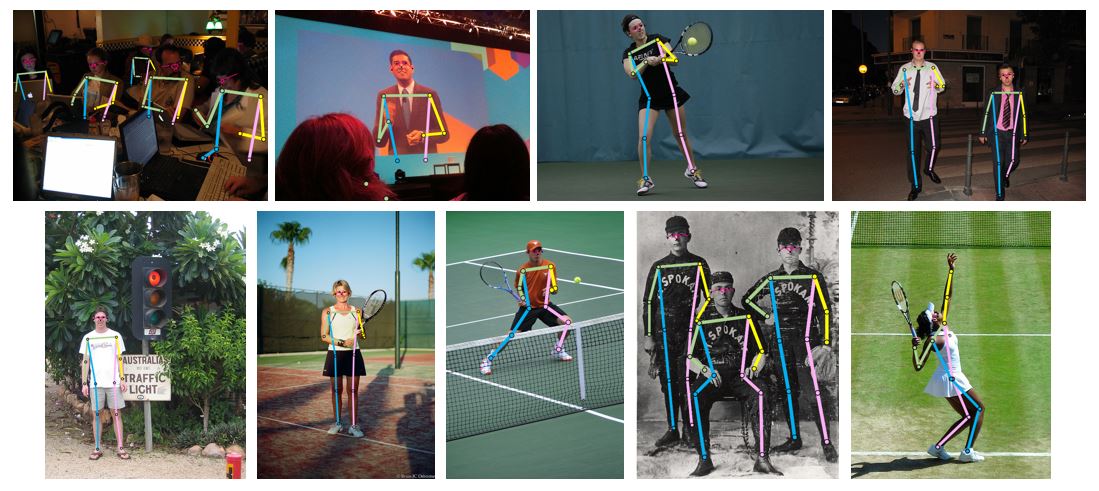} 
  \caption{Visualization of the pose estimation results based on AggPose-L on COCO val.}
  \label{fig:coco_vis}
\end{figure}

\begin{figure}[h]
  \centering
  \includegraphics[width=1.0\linewidth]{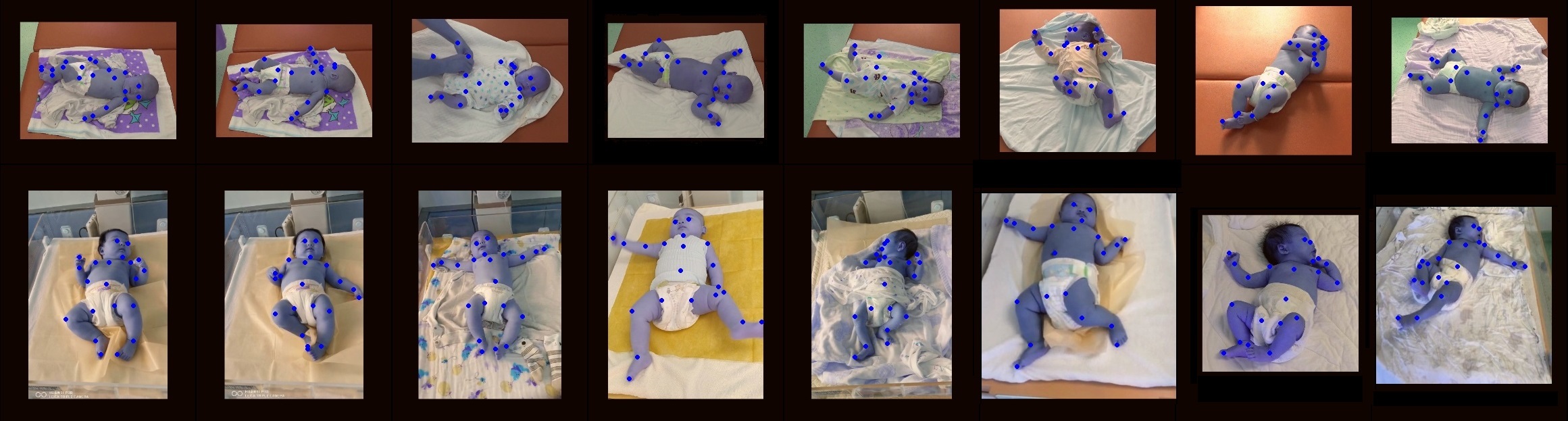} 
  \caption{Visualization of the pose estimation results based on AggPose-L on infant pose test.}
  \label{fig:infant_vis}
\end{figure}

\subsection{Ablation Experiments}

In previous sections, we compare AggPose with several state-of-art human pose estimation methods. To verify the techniques used in our method, we make detailed ablation studies in this subsection.

\paragraph{Influence of full Transformer backbone.} Considering all of the other new proposed hybrid methods are using HRNet's CNN encoder as backbone, we compare our method (without using convolution layers in the first stage) with the CNN scheme of HRNet in Table~\ref{tab:COCO_comparison}. The Backbone column shows the difference of the first stage inside the model. Both AggPose-S and AggPose-L are using SegFormer, a Transformer-based method as the first stage layer. Although other authors claim Transformers in the early stage will lead to lack detailed localization information, we observe that the full Transformer-based early stage of AggPose can still achieve better performance.

\paragraph{Influence of Deep Aggregation Framework.} We report the COCO pose estimation results based on two well-known full transformer models, Swin Transformer and SegFormer in Table~\ref{tab:abulation_comparison}. Both the Swin-B and SegFormer-B5 are pre-trained on ImageNet21K and fine-tuned on COCO with 300 epochs. In fact, AggPose-L can be considered as a deep layer aggregation structure of SegFormer-B5 with MLP skip-connection. According to the results in Table~\ref{tab:abulation_comparison}, our proposed multi-resolution aggregation framework (AggPose) achieves better performance than both Swin Transformer and SegFormer.

\subsection{Visualization Analysis}
We provide qualitative results on both COCO val set and infant pose test set, as shown in Figure~\ref{fig:heatmap_infant}, Figure~\ref{fig:coco_vis} and Figure~\ref{fig:infant_vis}. Figure~\ref{fig:heatmap_infant} shows a group of predictions and dependency areas for infant pose heatmap. Although infant pose data formats use more keypoints than COCO. AggPose still learns good representations in capturing constraint relationships between human keypoints.

\section{Conclusion}
By leveraging a new dataset with pose labels and clinical labels, we built a Transformer-based infant pose estimation framework which can accurately detect infant supine position pose from movement frames in video. The key insight of the AggPose model is the deep aggregation Transformer with cross-layer MLP connection. The pose sequence generated by our model has been used in neurodevelopmental disorder prediction for newborns and early evaluation for related diseases. Besides, our method can be packaged to mobile devices in the future and solve inequality in medical resources and privacy protection for patients. Although our results are promising, we acknowledge that there is still a long path to apply our model completely end-to-end with currently available hardware.

In addition to testing the utility of AggPose in real-time infant pose extraction and evaluation, a clear next step would be predicting the cerebral palsy via pose sequence understanding models in the future–before it is even visible to a trained pediatrician eye. For countries with limited pediatricians, this will greatly reduce the risk of severe disability in children.

\section*{Acknowledgments}

This work was supported by Sanming Project of Medicine in Shenzhen, China (SZSM202011005).

\bibliographystyle{named}
\bibliography{ijcai22}

\end{document}